\documentclass{article} 
\usepackage{iclr2019_conference,times}

\usepackage{amsmath,amsfonts,bm}

\def\eqref#1{equation~\ref{#1}}

\def\1{\bm{1}}

\DeclareMathAlphabet{\mathsfit}{\encodingdefault}{\sfdefault}{m}{sl}
\SetMathAlphabet{\mathsfit}{bold}{\encodingdefault}{\sfdefault}{bx}{n}

\usepackage{hyperref}
\usepackage{url}
\usepackage{graphicx}
\usepackage{subcaption}

\title{On the effectiveness of task granularity for transfer learning}

\author{Farzaneh Mahdisoltani \& David Fleet \\
Department of Computer Science,\\
University of Toronto,\\
Vector Institute. \\
\texttt{\{farzaneh, fleet\}@cs.toronto.edu} \\
\AND
Guillaume Berger \& Waseem Gharbieh \& Roland Memisevic\\
Twenty Billion Neurons Inc.\\
\texttt{\{firstname.lastname\}@twentybn.com} \\
}

\iclrfinalcopy 
\begin{document}

\maketitle

\begin{abstract}
\vspace*{-0.2cm}
We describe a DNN for video classification and captioning, trained
end-to-end, with shared features, to solve tasks at different levels of 
granularity, exploring the link between granularity in a source 
task and the quality of learned features for transfer learning. 
For solving the new task domain in transfer learning, we freeze
the trained encoder and fine-tune a neural net on the target domain.
We train on the Something-Something dataset with over $220,000$ 
videos, and multiple levels of target granularity, including $50$ action 
groups, $174$ fine-grained action categories and captions.
Classification and captioning with Something-Something are challenging 
because of the subtle differences between actions, applied to thousands 
of different object classes, and the diversity of captions penned by crowd actors.
Our model performs better than existing classification baselines 
for Something-Something, with impressive fine-grained results.  And it yields
a strong baseline on the new Something-Something captioning task. 
Experiments reveal that training with more fine-grained tasks tends 
to produce better features for transfer learning.
\end{abstract}

\vspace*{-0.2cm}
\section{Introduction}

\vspace*{-0.2cm}
Common-sense video understanding entails fine-grained recognition of actions, 
objects, spatial temporal relations, and physical interactions, 
arguably well beyond the capabilities of current techniques. 
A general framework will need to discriminate myriad variations of actions 
and interactions.
For example, we need to be able to discriminate similar 
actions that differ in  subtle ways, e.g., {\em 'putting a pen beside the cup'}, 
{\em 'putting the pen in the cup'}, or {\em 'pretending to put the pen on the 
table'}. Similarly, one must be able to cope with diverse actors and object classes. 
Such generality and fine-grained discrimination are key challenges to video 
understanding. In contrast to current approaches to action recognition and 
captioning on relatively small corpora, with coarse-grained actions, this paper 
considers fine-grained classification and captioning tasks on large-scale video 
corpora. Training is performed on the Something-Something dataset (\cite{GoyalICCV17}), 
with $174$ fine-grained action categories and thousands of different  objects, 
all under significant variations in lighting, viewpoint, 
background clutter, and occlusion. 

We describe a DNN architecture comprising a 2-channel convolutional net and an LSTM 
recurrent network for video encoding. A common encoding is shared for classification 
and caption generation.  The resulting network performs better than baseline action 
classification  results in \cite{GoyalICCV17}, and it gives impressive results on an 
extremely challenging fine-grained captioning task. 
We also demonstrate the quality of learned features through transfer learning from 
Something-Something features to a kitchen action dataset.
The main contributions in the paper include:

\noindent
1.\ \textbf{Explore the link between label granularity and feature quality}: We exploit 3 levels of granularity in sth-sth-v2, namely, action groups, action categories, and captions. Experiments show that more fine-grained labels yield richer features (see Table 3 and Fig.\ 4).

\noindent 
2.\ \textbf{Baselines for captioning on sth-sth-v2 data}: We note that the captioning task is new for this dataset; the original dataset did not provide captions.

\noindent
3.\ \textbf{Captioning as a source task for transfer learning}: We show that models trained to jointly perform classification and captioning learn features that transfer better to other tasks (e.g., see Fig.\ 4). To the best of our knowledge, captioning has, to date, been used as a target task. Our results suggest that captioning is a powerful source task.

\noindent
4.\ \textbf{20bn-kitchenware}: We introduce a new dataset, ostensibly for video transfer learning.

\vspace*{-0.2cm}
\section{Video Data}
\vspace*{-0.2cm}
Video action classification and captioning have received 
significant attention for several years, but progress has been
somewhat slow in part because of a lack of large-scale corpora. 
Using web sources (e.g., YouTube, Vimeo and Flickr) and human 
annotators, larger datasets have been collected
(e.g., \cite{KayEtAl2017,MonfortEtAl2018}), but they lack control over 
pose variations, motion and other scene properties that might be 
important for learning fine-grained models.
More recently, crowd-sourced/crowd-acted data has emerged.
This allows targeted video domains, action classes, and control 
over subtle differences between fine-grained action classes.
The first version of \cite{GoyalICCV17} has $100,000$ videos of 
human-object interactions, comprising $50$ coarse-grained action groups, 
decomposed further into $174$ related action categories. 
The videos exhibit significant diversity in viewing and lighting, 
objects and backgrounds, and the ways in which actors perform actions.
Baseline performance in \cite{GoyalICCV17} was a correct 
action classification rate of $11.5\%$, and $36.2\%$ on action groups. 
\cite{torralba_trn} report $42.01\%$ classification accuracy on 
Something-Something-V1 action categories. With our architecture we 
obtain validation accuracy of 38.8\%.

Something-Something-V2 is larger, with $220,847$
videos of the same $174$ action categories. In addition, each video
includes a caption that was authored and uploaded by the crowd actor. 
These captions incorporate the action class as well as descriptions 
of the objects involved. That is, the captions mirror 
the action template, but with the generic placeholder 
{\em Something} replaced by the object(s) chosen by 
the actor. As an example, a video with template action {\em 'Putting 
[something] on [something]'}, might have a caption {\em 'Putting a 
red cup on a blue plastic box'.}
In a nutshell, this dataset provides different levels of granularity:
$50$ coarse-grained action groups, $174$ fine-grained action categories,
and even more fine-grained labeling via video captions. 
\vspace*{-0.2cm}
\section{Video Classification and Captioning Tasks}

\vspace*{-0.2cm}
Due to the prevalence of datasets like UCF-101 (\cite{UCF101}), 
sports1M (\cite{KarpathyCVPR2014}) and more recently Kinetics
(\cite{KayEtAl2017}), most research on action classification 
has focused on models for coarse-grained action classification. 
In extreme cases, action classes can be can be discriminated from 
a glimpse of the scene, often encoded in isolated frames; e.g., 
inferring ``soccer play'' from the presence of a green field.
Even when motion is essential to the action, many existing approaches 
do well by encoding rough statistics over velocities, directions, and 
motion positions. 
Little work has been devoted to the task of representing details 
of object interactions or how configurations change over time. 

Image and video captioning have received increasing attention since the 
release of captioning data, notably, Microsoft COCO (\cite{mscoco_cap}) 
and  MSR-VTT (\cite{xu2016msr}). 
One problem with existing captioning approaches is that many 
datasets implicitly allow models 
to ``cheat'', e.g., by generating phrases that are grammatically 
and semantically coherent, but only loosely related to the
fine-grained visual structure. 
It has been shown, for example, that a language model trained on 
unordered, individual words (e.g., object-nouns) predicted by a 
separate NN can compete with captioning model trained on the actual
task end-to-end (e.g., \cite{yao2015oracle,heuer2016generating}). 
Similarly, nearest neighbor methods have been surprisingly  
effective (\cite{NNcaptioning}). 

Captioning tasks, if designed appropriately, should capture detailed 
scene properties.  Labels with more subtle and fine-grained distinctions 
would directly expose the ability (or inability) of a network to 
correctly infer the scene properties encoded in the captions. The captions 
provided with the Something-Something dataset are  designed to be sufficiently 
fine-grained that attention to details is needed to yield high prediction 
accuracy.   For example, to generate correct captions, networks need not 
only infer the correct actions, 
but must also learn to recognize different objects and their 
properties, as well as object interactions. 
\vspace*{-0.2cm}
\section{Approach} 

\begin{figure}[t]
\vspace*{-0.8cm}
    \centering
    \includegraphics[width=11cm]{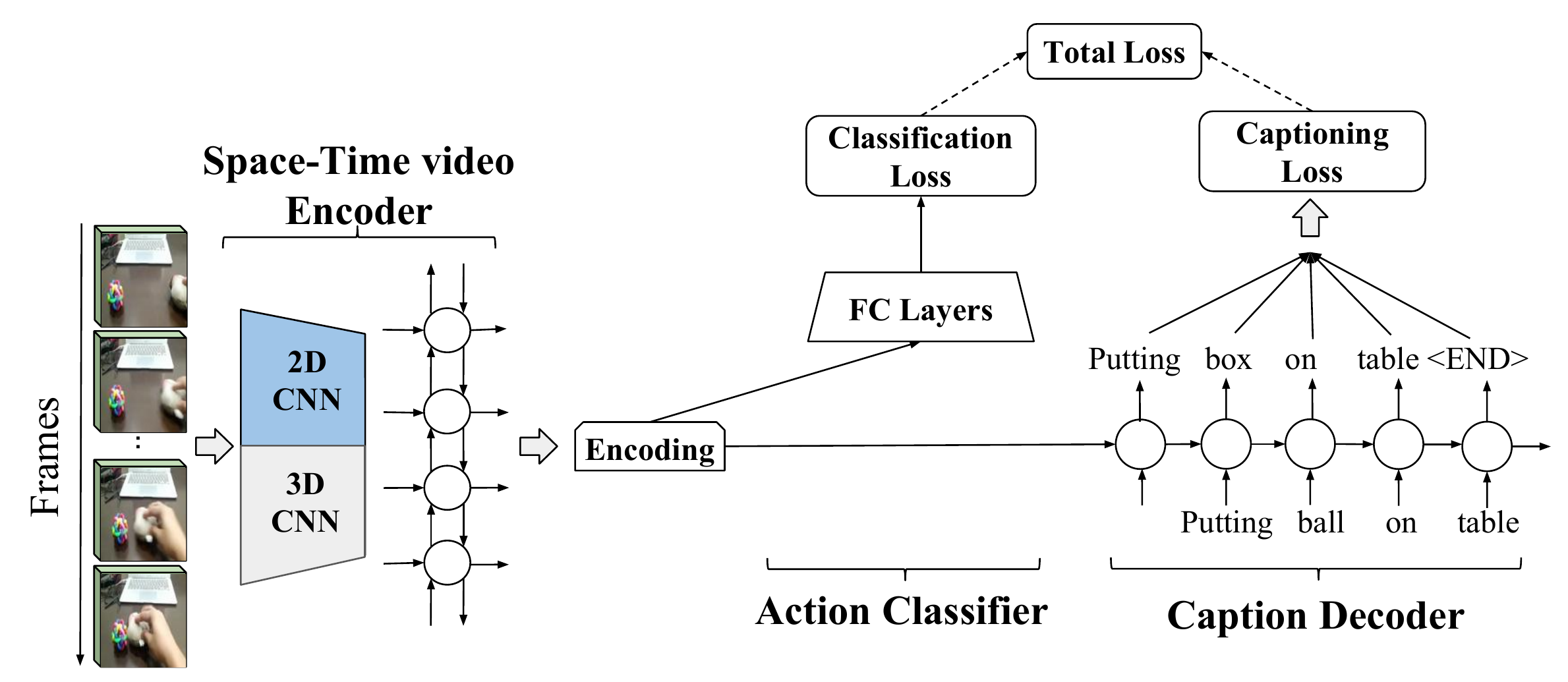}
    
\vspace*{-0.1cm}
    \caption{Our model architecture includes a two-channel CNN followed by
    an LSTM video encoder, an action classifier, and an LSTM decoder 
    for caption generation.}
    \label{fig:arch}
\vspace*{-0.2cm}
\end{figure}

\vspace*{-0.2cm}
We use a modified encoder-decoder architecture with an action classifier
applied to the video encoding (see Fig.\ \ref{fig:arch}). The decoder or 
classifier can be switched off, leaving pure classification or captioning 
models. One can also jointly train classification and captioning models. 
We train our two-channel architecture to solve four different tasks with different granularity levels:\\
\hspace*{0.25cm} - Coarse-grained classification on action groups (CG actions)\\
\hspace*{0.25cm} - Fine-grained classification on 174 action categories (FG actions)\\
\hspace*{0.25cm} - Captioning on simplified objects (SO captions)\\
\hspace*{0.25cm} - Fine-grained Captioning with full object placeholders (Captions)\\
Finally, we investigate the effectiveness of the learned features for transfer learning.

\begin{figure}[t]
    \centering
    \includegraphics[width=11cm]{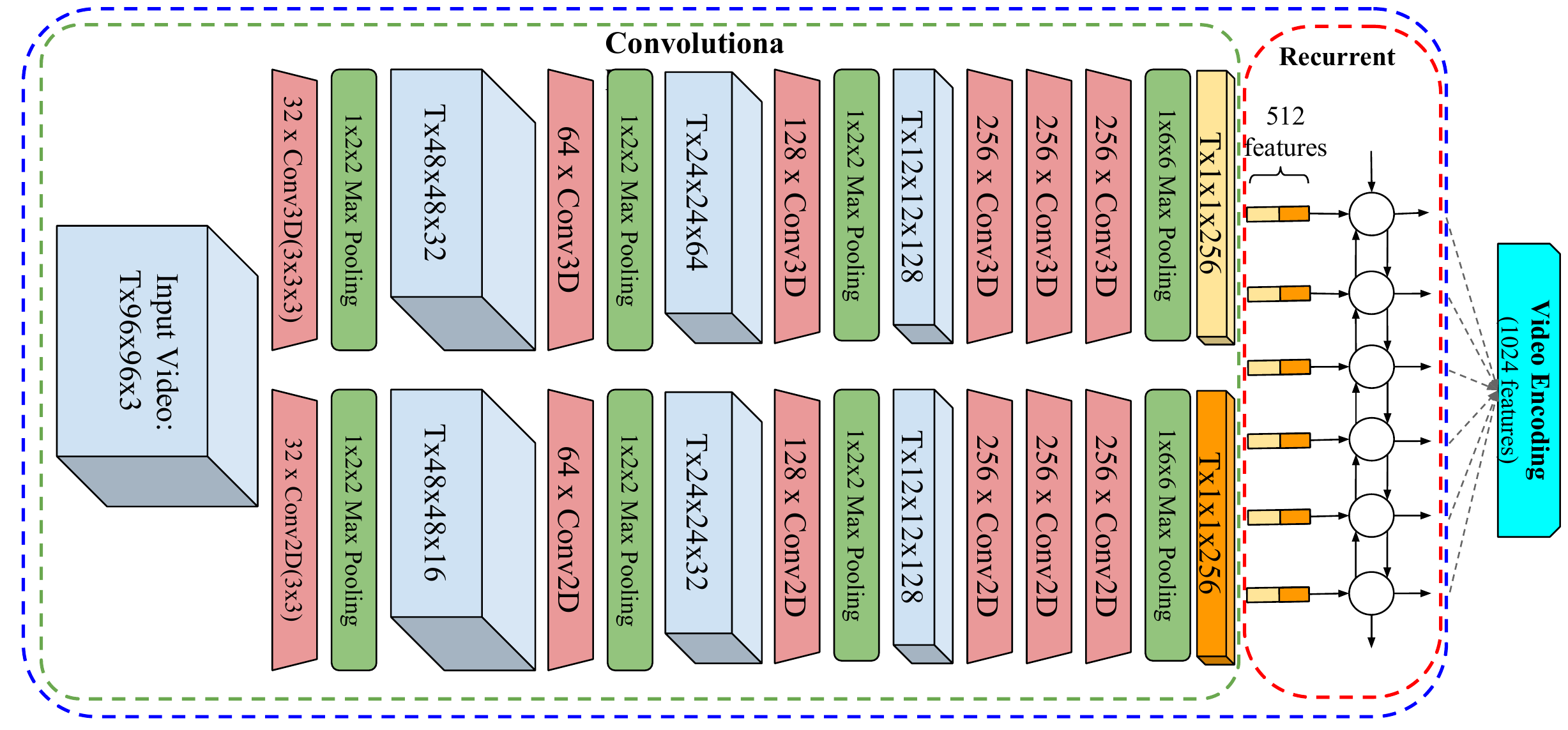}
    
\vspace*{-0.1cm}
    \caption{Our encoder includes a two-channel CNN followed by
    an LSTM for aggregating features.}
    \label{fig:encoder}
\end{figure}

\subsection{Modified Video Encoder-Decoder}

\vspace*{-0.2cm}
The {\em video encoder} receives the input video $v$, and maps 
it to an embedded representation $h$. Conditioned on $h$, a caption
decoder generates the caption $c$, and a classifier predicts the
action category. 
The encoder processes the video with a two-channel 
convolutional architecture (Fig.\ \ref{fig:encoder}). 
A spatial 2D-CNN and a spatiotemporal 3D-CNN are applied in 
parallel. 
The 3D-CNN features are used in lieu of a 
separate module to compute motion (e.g., optical flow) features.
The basic building block of each channel is a $3\times3\times3$ ($3\times3$ 
in 2D-CNN channel) convolution filter with batchnorm and ReLU activation. 
To aggregate features across time, feature vectors from 
each channel are combined and fed to a 2-layer bidirectional LSTM. 
We average these features to get the video encoding, $h$. 

The {\em action classifier} applies a fully-connected layer to the 
encoding $h$, followed by a softmax layer. Training uses a cross-entropy 
loss over action categories. The {\em caption decoder} is a two-layer LSTM. 
Much like conventional encoder-decoder methods for video captioning 
(\cite{venugopalan2014translating,donahue2015long, tesselation} and \cite{MT}), 
our decoder generates captions using a softmax over  vocabulary words, 
conditioned on previously generated words. The loss is the usual negative 
log-probability of the word sequence:  
\begin{equation}
     \mathrm{loss}_{\mathrm{captioning}} = -\sum_{i=0}^{N-1} \log p(w^{i+1}|w^{\le i}, h ;\theta).
\end{equation}
where $w^i$ denotes the $i^\text{th}$ word of the caption, $h$ is 
the video encoding, and $\theta$ denotes model parameters.
To optimize speed and memory usage during training, 
the length of captions generated by the decoder is fixed at
14 words\footnote{Less than $1\%$ of Something-Something 
captions have more than 14 words.}. As is common for encoder-decoder 
networks, we train with teacher-forcing (\cite{WilliamsZipser89}). 
At test time, the input to the decoder at each time-step is the token 
generated at the previous time-step (i.e., no teacher forcing).
The model is trained end-to-end for classification and captioning
with a weighted sum of the classification and captioning losses, i.e.,
\begin{equation}
    \mathrm{loss} = \lambda \cdot {\mathrm{loss}}_{\mathrm{classification}} + (1-\lambda) \cdot {\mathrm{loss}}_{\mathrm{captioning}}
\end{equation}
With $\lambda=1$ or $\lambda=0$, we end up with pure classification and 
captioning tasks respectively. For other values of $\lambda$, they are
trained jointly. The encoder is shared by the action classifier
and the caption decoder. The experiments below also compare this 
joint training regime with models for which the encoder is trained
on the classification loss or the captioning loss alone.

\vspace*{-0.2cm}
\paragraph{Related work:} Existing action classification methods
differ in the way they aggregate information over time. Many approaches 
rely  on spatial features with CNNs applied to individual frames (e.g., see 
\cite{KarpathyCVPR2014}). While such features have been successful, they fail to 
exploit temporal information in video.  
Other approaches  use spatial and  temporal  information 
(e.g., \cite{KarpathyCVPR2014,tran2015learning,3DCNN}).
Recent methods involve Imagenet-pretrained 2D-CNNs that are inflated into 3D (\cite{carreira2017quo,HaraCVPR2018}). 
Our video encoder is most closely related to approaches
that perform temporal reasoning via a recurrent convolutional architecture (\cite{molchanov2016online,baccouche2011sequential,li2017videolstm}). 
It is also related to TwoStream architectures (\cite{simonyan2014two});
but our model uses generic 3D CNN features rather than optical flow.

\vspace*{-0.2cm}
\section{Experiments} 

\vspace*{-0.2cm}
We train four two-channel 
models that are trained to solve coarse-grained and fine-grained 
classification and captioning tasks. In what follows we discuss different tasks in more details.

\subsection{Coarse- and Fine-Grained Classification}
\vspace*{-0.2cm}
Something-Something provides coarse-grained categories (action groups), each 
comprising a set of fine-grained actions. 
We trained a classification model on coarse-grained action groups,
using the M($256$-$256$) architecture, with accuracy of $57.6\%$ 
(see Table \ref{table:fg-cg} (top-left)).
Table  \ref{table:classif} reports the performance of our model for 
fine-grained classification.
For the pure classification task (with $\lambda=1$) 
we consider different numbers of features produced by the 
2D (spatial) CNN and the 3D (spatio-temporal) CNN. The total number of 
features is $512$ in all cases. 
The results show that the model benefits from  2D and 3D features. 
The even split M($256$-$256$) 
provides a good trade-off between performance and model complexity. 
We therefore use this model below, unless otherwise specified.
\begin{table}[t]{\footnotesize}
\vspace*{-0.5cm}
\centering
\begin{tabular}{l|c|c|c|c|c}
{\bf Models} &
 {\bf\begin{tabular}[c]{@{}l@{}}3D-CNN\\ Channels\end{tabular}}&
  {\bf\begin{tabular}[c]{@{}l@{}}2D-CNN\\ Channels\end{tabular}}&
  {\bf \begin{tabular}[c]{@{}l@{}}Number of\\ Parameters\end{tabular}}&
  {\bf\begin{tabular}[c]{@{}l@{}}Validation\\ Accuracy\end{tabular}} & {\bf\begin{tabular}[c]{@{}l@{}}  Test\\ Accuracy\end{tabular}} \\  \hline \hline
   M($256-0$) & 256 & 0 & 8.9M & 50.06 & 48.84 \\ \hline
   M($512-0$)&512 & 0 & 24.1M & 51.96 & {\bf51.15} \\ \hline
   M($384-128$)&384 & 128 & 16.2M & 51.11 & 49.96 \\ \hline
   M($256-256$)&256 & 256 & 11.5M & 51.62 & {\bf50.44} \\ \hline
   M($128-384$)&128 & 384 & 10.0M & 50.82 & 49.57 \\ \hline
   M($0-512$)&0 & 512 & 5.8M & 39.78& 37.80\\ \hline
   M($0-256$)&0 & 256 & 11.5M & 40.2 & 38.83 \\  
\end{tabular}
\vspace*{-0.1cm}
\caption{Validation and test accuracy for the pure classification task 
$(\lambda = 1)$,  with different numbers of 2D-CNN and 3D-CNN 
features used for video encoding.}
\label{table:classif}
\vspace*{-0.2cm}
\end{table}

\vspace*{-0.5cm}
\paragraph{Transferring between Coarse- and Fine-Grained Classification}
We can perform coarse-grained classification by
mapping predictions from the fine-grained classifier onto the action
groups. To this end we sum the probabilities of fine-grained actions 
belonging to each action group.  Interestingly, 
the resulting accuracy on coarse-grained action groups increases
to $62.5\%$. This improvement suggests that
fine-trained training provides higher quality features.
We also examine to what extent coarse-grained
performance accounts for fine-grained accuracy; i.e, how
better fine-grained performs compared to chance when conditioned on 
coarse-grain predictions.  For example, conditioned on a predicted 
action group, if we select the most frequent action within the 
action group, fine-grained test accuracy would be
$23.8\%$. One can also fix the coarse-grained model and 
train a linear classifier on top of its penultimate features. This
yields test accuracy of $41.7\%$, still $8.7\%$ below test
performance for the corresponding architecture trained on the 
fine-grained task, further supporting our contention that training 
on fine-grained details yields richer features.

\begin{table}[t!]{\footnotesize}
\vspace*{-0.2cm}
\centering
\begin{tabular}{l|c|c}
 & \begin{tabular}[c]{@{}l@{}}Coarse-grained Testing \end{tabular} & 
  \begin{tabular}[c]{@{}l@{}}Fine-grained Testing \end{tabular}\\ \hline \hline
Coarse-grained Training & 57.6 & 41.7 \\ \hline
Fine-grained Training& {\bf 62.5} & {\bf50.44} \\ 
\end{tabular}
\caption{Comparison of classification accuracy of fine-grained and 
coarse-grained models, tested on fine-grained actions (using action categories) versus coarse-grained actions (using action groups).}
\label{table:fg-cg}
\vspace*{-0.2cm}
\end{table}
\vspace*{-0.2cm}
\subsection{Classification Baselines.}
As a baseline, we use ImageNet-pretrained models (\cite{vgg_simonyan,resnet})
on individual frames, to which we add  layers. 
First, we use just the middle frame of the video, with
a 2-layer MLP with $1024$ hidden units. We also 
consider a baseline in which we apply this approach to all $48$ frames, 
and then  average frame by frame predictions.
We experiment with aggregating information over time using a LSTM 
layer with 1024 units. We report results in Table \ref{table:baseline}.  
There is a marked improvement with the LSTM, confirming that this task 
improves with temporal analysis.  Our best baseline  
was achieved with a VGG architecture, and the test accuracy is close 
to the best architecture reported to date on Something-Something
(e.g., \cite{torralba_trn}).
\begin{table}[t]{\footnotesize}
\centering
\begin{tabular}{l|c}
{\bf Models}  & {\bf \begin{tabular}[c]{@{}l@{}}Test Accuracy\end{tabular}} \\ \hline \hline
\begin{tabular}[c]{@{}l@{}}VGG16 + MLP 1024 (single middle frame)\end{tabular} & 13.29 \\ \hline
\begin{tabular}[c]{@{}l@{}}VGG16 + MLP 1024 (averaged over 48 frames)\end{tabular} &  17.57 \\ \hline
\begin{tabular}[c]{@{}l@{}}VGG16 + LSTM 1024\end{tabular}  & {\bf31.69} \\ \hline
\begin{tabular}[c]{@{}l@{}}ResNet152 + MLP 1024  (single middle frame)\end{tabular}  & 13.62 \\  \hline
\begin{tabular}[c]{@{}l@{}}ResNet152 + MLP 1024  (averaged over 48 frames ) \ \ \end{tabular}  & 16.79 \\ \hline
\begin{tabular}[c]{@{}l@{}}ResNet152 + LSTM 1024 (48 steps)\end{tabular}  & {\bf28.82} \\ 
\end{tabular}
\vspace*{-0.2cm}
\caption{Classification results on 174 action categories using VGG16 and 
ResNet152 as frame encoders, along with different strategies for 
temporal aggregation. 
}
\label{table:baseline}
\vspace*{-0.3cm}
\end{table}

\subsection{Captioning with simplified object placeholders.}

\vspace*{-0.2cm}
The ground truth object placeholders in Something-Something video captions 
(i.e., the object descriptions provided by crowd actors) are not highly 
constrained. Crowd actors have the option to type in the objects they 
used, along with multiple descriptive or quantitative adjectives, 
elaborating shape, color, material or the number of objects involved.
Accordingly, it is not surprising that the distribution over object 
placeholders is extremely heavy-tailed, with many words occurring rarely. 
To facilitate training we therefore replaced all words that occurred $5$ 
times or less by [Something]. After removing rare words, we are left 
with $2880$ words comprising around $30,000$ distinct object placeholders
(i.e., different combination of nouns, adjectives, etc). 

We consider a simplified task in which we modify the ground truth 
captions so they only contain one word per placeholder, by keeping the last noun, 
removing all other words from the placeholders. By substituting the pre-processed
placeholders into the action category, we obtain a simplified 
caption. Table \ref{table:annot} shows an example of the process. 
The result is a reduced vocabulary with $2055$ words. In the spectrum 
of granularity, captioning with simplified objects can be considered as 
a middle ground between fine-grained action classification and captioning 
with full labels.

We train different variations of our two-channel architecture on captions with 
simplified objects. Table \ref{table:single_object_cap} summarizes our results. 
We observe that the model with an equal number of 2D- and 3D-channels 
performs best (albeit by a fairly small margin). Also the 
best captioning model performs best on the classification task. 
We also evaluate the models using standard captioning metrics: 
BLEU (\cite{papineni2002bleu}), ROUGE-L (\cite{lin2004rouge}) and METEOR (\cite{michael2014meteor}). 

\subsubsection{Fine-grained Captioning with full object placeholders.}

\vspace*{-0.2cm}
We also train networks on the full object placeholders. 
This constitutes the finest level of action granularity. 
The experimental results are shown in Table \ref{table:full_cap}. 
They show that, again, the best captioning model also yields the highest 
corresponding classification accuracy. The Exact-Match accuracy is 
significantly lower than for the simplified object placeholders, 
as it has to account for a much wider variety of phrases. 
The captioning models produce impressive qualitative results with a 
high degree of approximate action and object accuracy. 
Some examples are shown in Figure \ref{captioning_examples}. 
More examples can be found in the appendix. 

To the best of our knowledge there are no baselines for the 
Something-Something captioning task.
To quantify the performance of our captioning models, we 
count the percentage of generated captions that match ground truth 
word by word. We refer to this as ``Exact-Match Accuracy''.  
This is a challenging metric as the model is deemed correct only 
if it generates the entire caption correctly.
If we use the action category predicted by model M($256$-$256$), trained 
for classification, and replace all occurrences of [something] with the most 
likely object string conditioned on that action class, the Exact-Match 
accuracy is $3.15\%$. The same baseline for simplified object placeholders 
is $5.69\%$. We also implemented a conventional encoder-decoder model for captioning \ref{table:cap_baselines}.
\begin{table}[h]{\footnotesize}
\centering
\begin{tabular}{l|c|c|c|c|c}
{\bf Models} &
  {\bf\begin{tabular}[c]{@{}l@{}}BLEU@4\end{tabular}} &
  {\bf\begin{tabular}[c]{@{}l@{}}ROUGE-L\end{tabular}} &
  {\bf\begin{tabular}[c]{@{}l@{}}METEOR\end{tabular}} &
  {\bf\begin{tabular}[c]{@{}l@{}}Exact-Match\\Accuracy \end{tabular}} &
   {\bf\begin{tabular}[c]{@{}l@{}}Classification\\ Accuracy\end{tabular}} \\  \hline \hline
   VGG16+LSTM      & 31.83 & 52.22 & 24.79 & 3.13 & 31.69 \\ \hline
   Resnet152+LSTM  & 31.93  & 51.76  & 24.89  & 3.25 &28.82
\end{tabular}
\vspace*{-0.2cm}
\caption{Captioning baselines using a conventional encoder-decoder architecture}
\label{table:cap_baselines}
\vspace{-0.3cm}
\end{table}

\begin{table}[t]{\footnotesize}
\vspace*{-0.3cm}
\centering
\begin{tabular}{|l|l|}
\hline
{\bf Video ID} & 81955 \\ \hline 
{\bf Action Group} & Holding {[}something{]} \\ \hline 
{\bf Action Category} & Holding {[}something{]} in front of {[}something{]} \\ \hline
{\bf Somethings} & ``a blue plastic cap'', ``a men's short sleeve shirt'' \\ \hline
{\bf Simplified somethings's} & ``cap'', ``shirt'' \\ \hline
{\bf Simplified-object Caption} & Holding cap in front of shirt \\ \hline
{\bf Full Caption} & Holding a blue plastic cap in front of a men short sleeve shirt\\ \hline
\end{tabular}
\vspace*{-0.1cm}
\caption{An example of annotation file for a Something-Something video}
\label{table:annot}
\vspace{-1\baselineskip}
\end{table}
\vspace*{-0.4cm}
\paragraph{Training settings} In all our experiments we use frame 
rate of $12 fps$. During training we randomly pick $48$ consecutive 
frames. For videos with less than $48$ frames, we replicate the 
first and last frames to achieve the intended length. We resize 
the frames to $128\times128$, and then use random cropping of size 
$96\times96$. For validation and testing, we use $96\times96$ 
center cropping. We optimize all models using Adam, with 
an initial learning rate of $0.001$. 

\begin{table}[t]{\footnotesize}
\centering
\begin{tabular}{l|c|c|c|c|c}
{\bf Models} &

  {\bf\begin{tabular}[c]{@{}l@{}}BLEU@4\end{tabular}} &
  {\bf\begin{tabular}[c]{@{}l@{}}ROUGE-L\end{tabular}} &
  {\bf\begin{tabular}[c]{@{}l@{}}METEOR\end{tabular}} &
  {\bf\begin{tabular}[c]{@{}l@{}}Exact-Match\\Accuracy \end{tabular}} &
   {\bf\begin{tabular}[c]{@{}l@{}}Classification\\ Accuracy\end{tabular}} \\  \hline \hline
   M($256-0$)  & 22.75 & 44.54 & 22.40 & 8.46 & 50.64 \\ \hline
   M($512-0$)  & 23.28  & 45.29  & 22.75  & 8.47 &50.96\\ \hline
   M($384-128$) & 23.02  & 44.86  & 22.58 & 8.53 &50.73\\ \hline
   M($256-256$) & 23.04  & 44.89  & 22.60  & {\bf 8.63} & {\bf51.38}\\ \hline
   M($128-384$) & 22.76 & 44.40 & 22.39 & 8.33& 50.04
\end{tabular}
\vspace*{-0.1cm}
\caption{Performance of our two-channel models with different sizes of channel features on for {\bf simplified objects}. For this task we use $(\lambda = 0.1)$. The maximum sequence length is $14$.
}
\label{table:single_object_cap}
\end{table}

While the captioning task theoretically entails action classification, 
we found that our two-channel networks optimized on the pure captioning 
task do not perform as well as models trained jointly on classification 
and captioning (see \ref{table:pure_vs_joint}). By coarsely tuning the  
hyper-parameter $\lambda$ empirically, we found $\lambda=0.1$ to work 
well and fix it at this value for the captioning experiments below.
More specifically, we first train with a pure classification 
loss, by setting $\lambda=1$, and subsequently introduce the captioning 
loss by gradually decreasing $\lambda$ to 0.1.

\begin{table}[t]{\footnotesize}
\vspace*{-0.5cm}
\centering
\begin{tabular}{l|c|c}
{\bf Models} &
 {\bf\begin{tabular}[c]{@{}l@{}}Classification Accuracy\end{tabular}}&
  {\bf\begin{tabular}[c]{@{}l@{}}Exact-Match Accuracy \end{tabular}} \\  \hline \hline
   $\lambda=0$ & 39.78  & 5.96\\ \hline
   $\lambda=0.1$ & {\bf 51.32} & {\bf 8.63}\\ \hline
\end{tabular}
\vspace*{-0.1cm}
\caption{Comparing models trained with pure captioning task vs joint 
captioning and classification. Results are shown for captioning with 
simplified object placeholders. The test classification accuracy for 
the pure captioning model was obtained by freezing the video encoder 
and training a linear regressor on top of penultimate features.}
\label{table:pure_vs_joint}
\end{table}


\begin{table}[t]{\footnotesize}
\vspace*{-0.2cm}
\centering
\begin{tabular}{l|c|c|c|c|c}
{\bf Models} &

  {\bf\begin{tabular}[c]{@{}l@{}}BLEU@4\end{tabular}} &
  {\bf\begin{tabular}[c]{@{}l@{}}ROUGE-L\end{tabular}} &
  {\bf\begin{tabular}[c]{@{}l@{}}METEOR\end{tabular}} &
  {\bf\begin{tabular}[c]{@{}l@{}}Exact Match\\Accuracy \end{tabular}} &
   {\bf\begin{tabular}[c]{@{}l@{}}Classification\\ Accuracy\end{tabular}} \\  \hline \hline
   M($256-0$)    & 16.87 & 40.03 & 19.13 & 3.33 & 50.48\\ \hline
   M($512-0$)    & 16.92 & 40.54 & 19.26 & 3.56 & 49.81\\ \hline
   M($384-128$) & 17.99 & 41.82 & 20.03 & {\bf3.80} & {\bf50.92}\\ \hline
   M($256-256$) & 17.61 & 41.28 & 19.69 & 3.76 & 50.56\\ \hline
   M($128-384$) & 16.80 & 39.98 & 19.11 & 3.61 & 49.24
\end{tabular}
\vspace*{-0.1cm}
\caption{Performance of captioning models with different sizes 
of channel features on full object placeholders. For this task 
we use $(\lambda = 0.1)$.  The maximum sequence length is $14$.
}
\label{table:full_cap}
\end{table}

\begin{figure}[t!]
\vspace*{-0.2cm}
\centering
\begin{subfigure}[b]{0.8\textwidth}
  \includegraphics[width=1\linewidth]{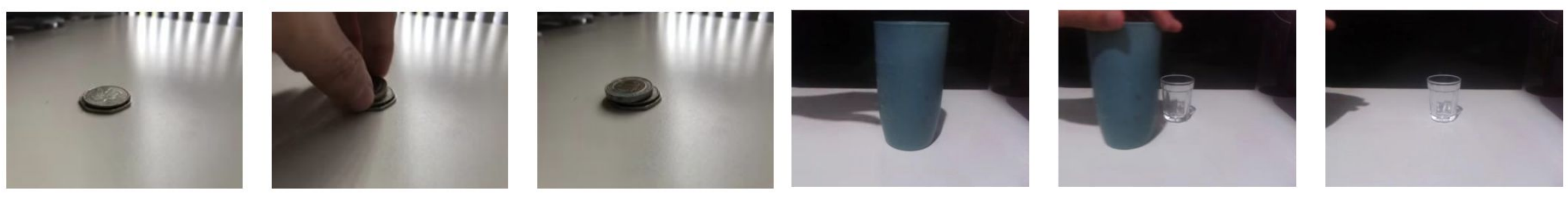}
  \vspace{-2\baselineskip}
  \label{kitchenware-3}
\end{subfigure}
\caption{Captioning examples: \\
    \textbf{Model Outputs:} [Piling coins up], [Removing mug, revealing cup behind].\\
    \textbf{Ground Truths:} [Stacking coins], [Removing cup, revealing little cup behind].}
\label{captioning_examples}
\vspace*{-0.3cm}
\end{figure}

   
\vspace*{-0.2cm}
\section{Transfer Learning}

\vspace*{-0.2cm}
One astonishing property of neural networks is their 
ability to learn representations transfer well to other tasks
(e.g., \cite{donahue2013decaf,sharif2014cnn}). 
A distinguishing feature of the ImageNet task, which likely contributes 
to its potential for transfer learning, is the dataset size and 
the  variety of fine-grained discriminations required. 
In what follows we explore transfer learning performance 
as a function of course task granularity.

\begin{figure}[t!]
\vspace*{-0.2cm}
\centering
\begin{subfigure}[b]{0.8\textwidth}
   \includegraphics[width=1\linewidth]{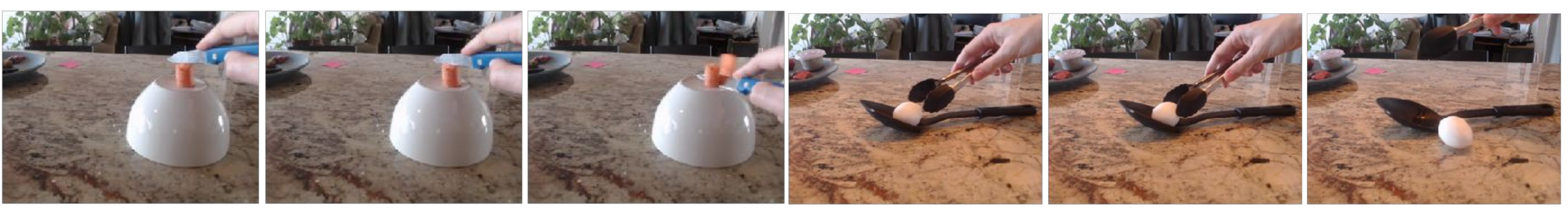}
   \vspace{-2\baselineskip}
   \label{kitchenware-3}
\end{subfigure}
\caption{20bn-kitchenware samples: Using a knife to cut something (left), Trying but failing to pick something up with tongs(right).}
\label{kitchenware-overview}
\vspace*{-0.1cm}
\end{figure}

We introduce \textit{20bn-kitchenware}, a few-shot video classification dataset with 
390 videos over 13 action categories (see Fig.\ \ref{kitchenware-overview}).
It contains video clips of manipulating a kitchen utensil for roughly 4 seconds and was
designed to capture fine-grained actions with subtle differences. 
For each  utensil $X \in \{\text{fork, spoon, knife, tongs}\}$, the target label belongs to one of 3 actions, {\em ``Using $X$"}, {\em ``Pretending to use $X$"} or {\em ``Trying but failing to use $X$"}. In addition to these 12 action categories, we also include a fall-back class labeled {\em ``Doing other things"}.

We further encourage the model to pay attention to visual details by including unused 
`negative' objects. The last row of Fig.\ \ref{kitchenware-overview} shows one example;  
the target label indicates a manipulation of tongs, but the clip also contains 
a spoon with an egg in it. Given the limited amount of data available for 
training\footnote{130 samples -- the rest are used for validation and testing.}, 
the action granularity and the presence of negative objects, we hypothesize that
only models that have some understanding of physical world properties will perform well. 
We will release 20bn-kitchenware upon publication of this paper.

\vspace*{-0.5cm}
\subsection{Proposed benchmark}


Our transfer learning experiments considers four two-channel 
models that have been respectively pre-trained on coarse-grained labels
(classification on action groups), on fine-grained labels (classification 
on 174 action categories), on simplified captions (captioning on 
fine-grained action categories expanded with single object descriptor)
and on template captions (captioning on fine-grained 
action categories expanded with object descriptors). 
We also include two neural nets  pre-trained on other datasets: a VGG16 network 
pre-trained on ImageNet, and an Inflated-ResNet34 pre-trained 
on Kinetics\footnote{https://github.com/kenshohara/3D-ResNets-PyTorch}.


The overall training procedure remains the same for all models. 
We freeze each pre-trained model and then train a neural net on top of extracted penultimate features.
Independent of the architecture used, we use the pre-trained model 
to produce $12$ feature vectors per second. To achieve this, where necessary we 
split the input video into smaller clips and apply the 
pre-trained network on each clip individually\footnote{VGG16 applied 
to individual frames. For Inflated-ResNet34, video was split into 
clips of 16 frames.}. In the simplest case, we pass 
the obtained features through a logistic regressor and
average predictions over time. We also report results for which we classify
pre-trained features using an MLP with 512 hidden units as well as
a single bidirectional LSTM layer with 128 hidden 
states. This allows the network to perform some temporal analysis about the 
target domain.

\begin{figure}[!t]
    \centering
    \includegraphics[width=0.8\textwidth]{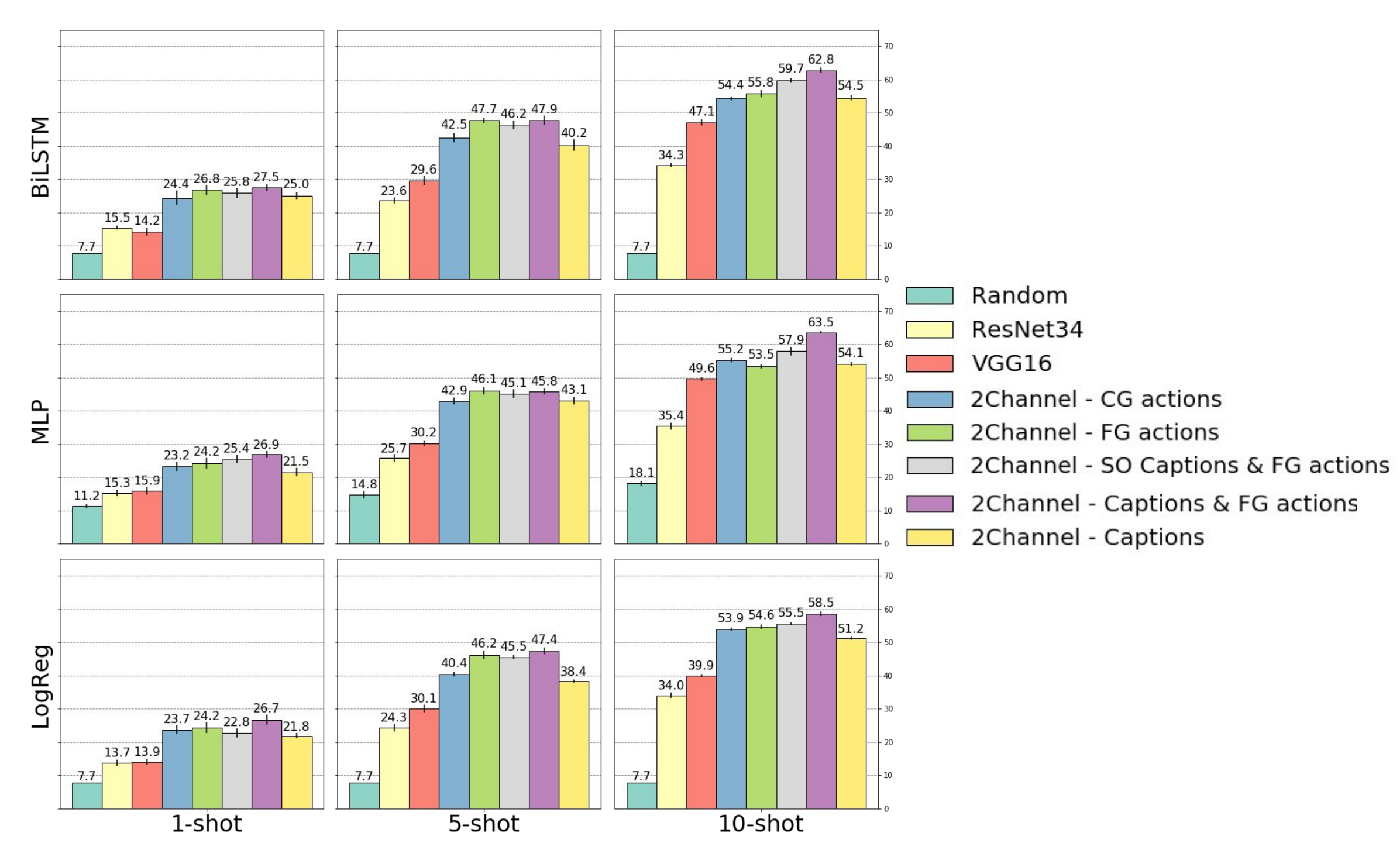}
    \caption{20bn-kitchenware transfer learning results: averaged scores 
    obtained using a VGG16, an Inflated ResNet34, 
    as well as two-channel models trained on four aforementioned tasks. 
    We report results using 1 training sample per class, 5 training 
    samples per class or the full training set.}
    \label{tl_results}
\end{figure}
\vspace*{-0.2cm}
\subsubsection{Observations}  
\vspace*{-0.2cm}
For each pretrained model and classifier, we evaluate 1-shot, 5-shot and 
10-shot performance, averaging scores over 10 runs. Fig.\ \ref{tl_results} 
shows the average scores with 95$\%$ confidence intervals. The most noticeable 
findings  are: 

\noindent
1.\ \textit{Logistic Regression vs MLP vs BiLSTM}:  Using a recurrent network yields better performance. 

\noindent
2.\ \textit{Something-Something features vs others}:  Our models pre-trained 
on Something-Something outperform other external models. This is not surprising 
given the target domain; 20bn-kitchenware samples are, by design, 
closer to Something-Something samples than ImageNet or Kinetics ones. It is surprising
that VGG16 features perform better  on 20bn-kitchenware than Kinetics features.

\noindent
3.\ \textit{Effect of the action granularity}: Fig.\ \ref{tl_results} supports 
the contention that training on fine-grained tasks yields better features. 
The best model on this benchmark is that trained jointly on full captions
and action categories. The only exception  is the 
model trained to do pure captioning.

    
    
\vspace*{-0.2cm}
\section{Conclusions}

\vspace*{-0.2cm}
Pre-training neural networks on large labeled datasets has become a 
driving force in deep learning applications. Some might argue that
it may be considered a serious competitor to unsupervised learning as 
a means to generate universal features for the visual world.
Ever since ImageNet became popular as a generic  feature extractor, a 
hypothesis been that it is dataset size, the amount of detail 
and the variety of labels that drive a network's capability to learn 
useful features. 
To the degree that this hypothesis is true, generating visual features 
capable of transfer learning should involve source tasks that (i) are 
fine-grained and complex, and (ii) involve video not still images, 
because video is a much more fertile domain for defining 
complex tasks that represent aspects of the physical world. 

This paper provides further evidence for that hypothesis, showing 
that task granularity has a strong influence on the 
quality of the learned features. We also show that captioning, which to 
the best of our knowledge has been used only as a \emph{target} 
task in transfer learning, can be a powerful source task. 
Our work suggests that one gets substantial leverage by utilizing ever 
more fine-grained recognition tasks, represented in the form of 
captions, possibly in combination with question-answering. 
Unlike the current trend of engineering neural networks to perform  
bounding box generation, semantic segmentation, or tracking, the appeal 
of fine-grained textual labels is that they provide a simple homogeneous
interface. More importantly, they may provide ``just enough'' localization 
and tracking capability to solve a wide variety of important tasks, without 
allocating valuable compute resources to satisfy intermediate goals at an 
accuracy that these tasks may not actually require. 

\bibliography{iclr2019_conference}
\bibliographystyle{iclr2019_conference}

\clearpage
\section*{Appendix}
In this supplementary document, we provide:
\begin{itemize}
    \item Qualitative examples for our classification and captioning.
    \item Visualization of our classification and captioning models using Grad-CAM.
    \item The full list of action categories of 20bn-kitchenware. 
\end{itemize}

\subsection*{Qualitative Examples of Classification}
\begin{figure}[!h]
\centering
\begin{subfigure}[b]{0.90\textwidth}
  \includegraphics[width=1\linewidth]{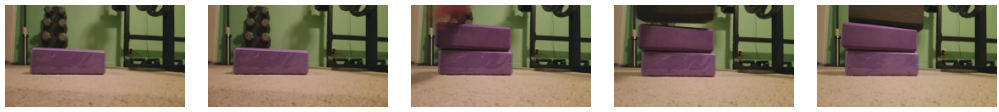}
  \vspace{-1.5\baselineskip}
  \caption{{\bf Ground Truth:} Stacking [number of] [something].}
  \vspace{-0.65\baselineskip}
  \caption{{\bf Model Prediction:} Putting [something] and [something] on the table.}
  \label{classif_ex100} 
\end{subfigure}
\begin{subfigure}[b]{0.90\textwidth}
  \includegraphics[width=1\linewidth]{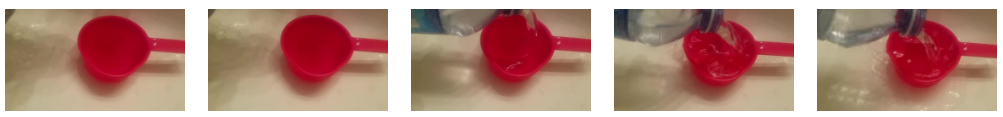}
  \vspace{-1.5\baselineskip}
  \caption{{\bf Ground Truth:} Pouring [something] into [something] until it overflows.}
  \vspace{-0.65\baselineskip}
  \caption{{\bf Model Prediction:} Pouring [something] into [something].}
  \vspace{0.4\baselineskip}
  \label{classif_ex100} 
\end{subfigure}

%
\begin{subfigure}[b]{0.90\textwidth}
  \includegraphics[width=1\linewidth]{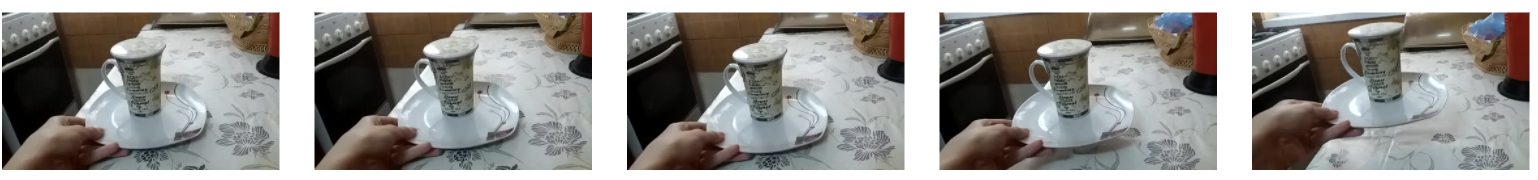}
  \vspace{-1.5\baselineskip}
  \caption{{\bf Ground Truth:} Lifting [something] with [something] on it.}
  \vspace{-0.65\baselineskip}
  \caption{{\bf Model Prediction:} Lifting [something] with [something] on it.}
  \vspace{0.4\baselineskip}
  \label{classif_ex100} 
\end{subfigure}

\begin{subfigure}[b]{0.90\textwidth}
  \includegraphics[width=1\linewidth]{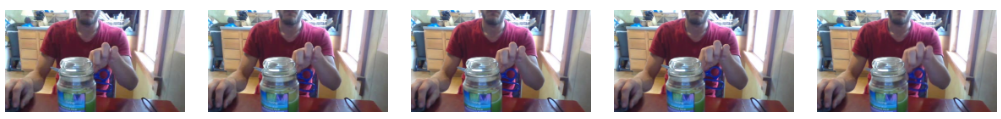}
  \vspace{-1.5\baselineskip}
  \caption{{\bf Ground Truth:} Showing [something] behind [something].}
  \vspace{-0.65\baselineskip}
  \caption{{\bf Model Prediction:} Holding [something] behind [something].}
  \vspace{0.4\baselineskip}
  \label{classif_ex100} 
\end{subfigure}

\begin{subfigure}[b]{0.90\textwidth}
  \includegraphics[width=1\linewidth]{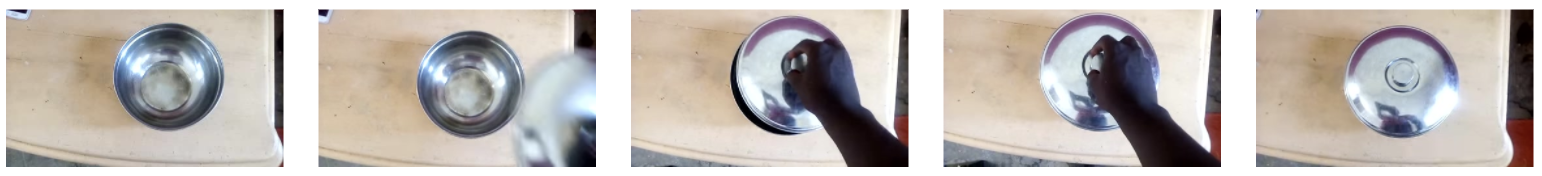}
  \vspace{-1.5\baselineskip}
  \caption{{\bf Ground Truth:}Putting [something] onto [something].}
        \vspace{-0.65\baselineskip}
  \caption{{\bf Model Prediction:}Covering [something] with [something].}
   
  \vspace{0.4\baselineskip}
  \label{classif_ex103} 
\end{subfigure}
\caption{Ground truth and model prediction for classification examples.}
\label{classif_examples}
\end{figure}

Here we provide video examples and their ground truth action categories, along with model predictions for each. We use our M($256$-$256$) which is trained 
with $\lambda=0.1$. Interestingly, notice that even when the predicted 
actions are incorrect, e.g. row $4$ in Figure \ref{classif_examples}, they
are, nevertheless, usually quite sensible.

\newpage

\subsection*{Qualitative Examples of Captioning}

Below are examples of videos. accompanied by their their ground truth 
caption and the caption generated by the model.  
We use model M($256$-$256$) in this section as well, which is also
trained jointly for classification and captioning (with $\lambda=0.1$).

\begin{figure}[!h]
\centering

\begin{subfigure}[b]{0.95\textwidth}
  \includegraphics[width=1\linewidth]{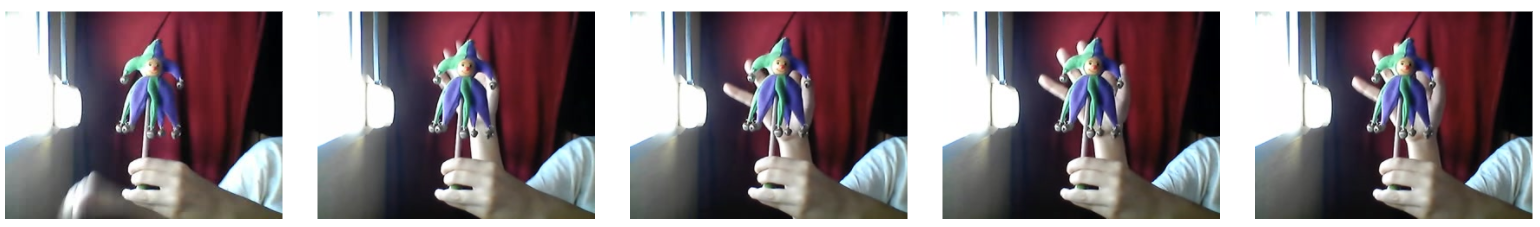}
  \vspace{-1.5\baselineskip}
  \caption{{\bf Ground Truth:} Touching (without moving) the head of a toy.}
  \vspace{-0.65\baselineskip}
  \caption{{\bf Model output:} Poking a stuffed animal so lightly that it doesnt or almost doesnt move.}
  \vspace{0.4\baselineskip}
  \label{cap_ex101} 
\end{subfigure}
\begin{subfigure}[b]{0.95\textwidth}
  \includegraphics[width=1\linewidth]{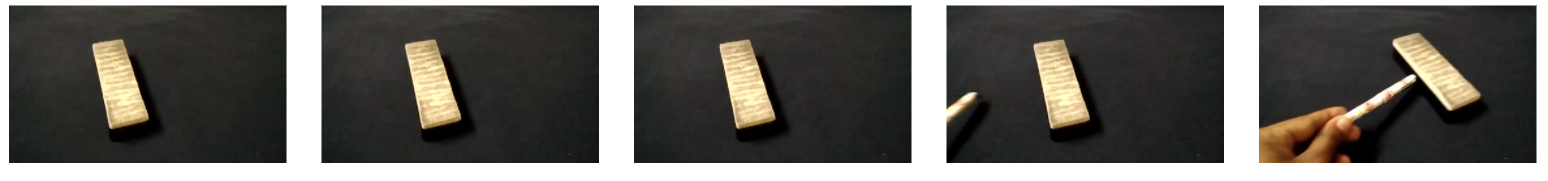}
  \vspace{-1.5\baselineskip}
  \caption{{\bf Ground Truth:}  Pushing duster with white coloured pen.}
  \vspace{-0.65\baselineskip}
  \caption{{\bf Model output:} Pushing phone with pen.}
  \vspace{0.4\baselineskip}
  \label{cap_ex101} 
\end{subfigure}
\begin{subfigure}[b]{0.95\textwidth}
  \includegraphics[width=1\linewidth]{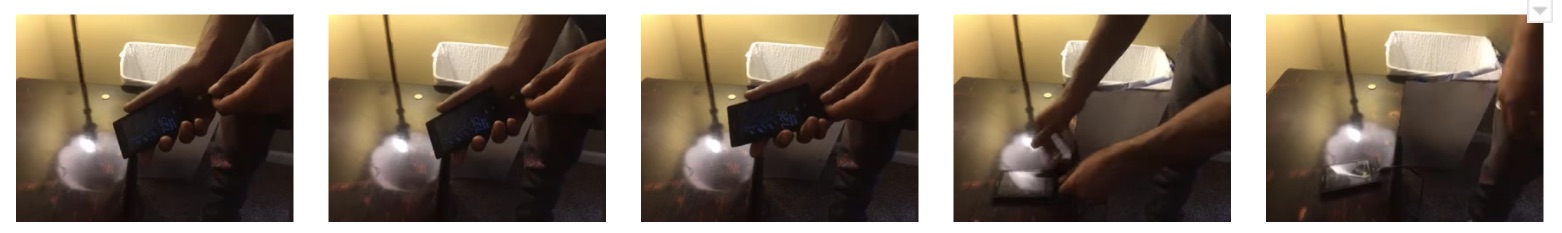}
  \vspace{-1.5\baselineskip}
  \caption{{\bf Ground Truth:} Plugging a charger into a phone.}
  \vspace{-0.65\baselineskip}
  \caption{{\bf Model output:} Plugging charger into phone.}
  \vspace{0.4\baselineskip}
  \label{cap_ex100} 
\end{subfigure}
\begin{subfigure}[b]{0.95\textwidth}
  \includegraphics[width=1\linewidth]{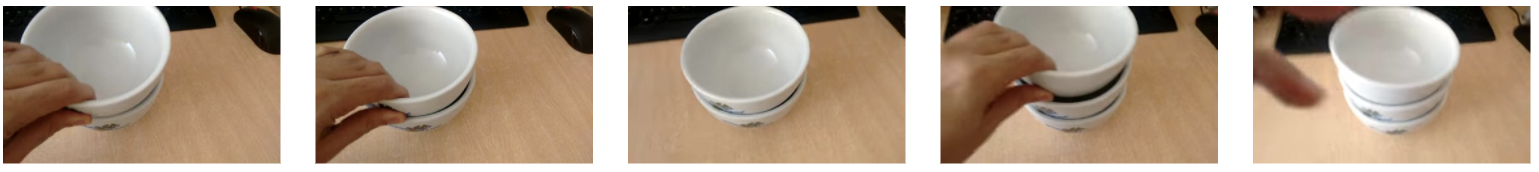}
  \vspace{-1.5\baselineskip}
  \caption{{\bf Ground Truth:} Piling bowl up.}
  \vspace{-0.65\baselineskip}
  \caption{{\bf Model output:} Stacking bowls.}
  \vspace{0.4\baselineskip}
  \label{cap_ex102} 
\end{subfigure}
\begin{subfigure}[b]{0.95\textwidth}
  \includegraphics[width=1\linewidth]{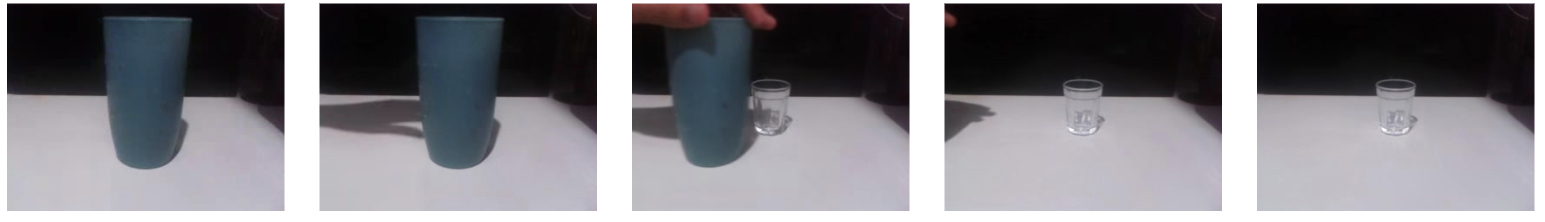}
  \vspace{-1.5\baselineskip}
  \caption{{\bf Ground Truth:} Removing cup, revealing little cup behind.}
  \vspace{-0.65\baselineskip}
  \caption{{\bf Model output:} Removing mug, revealing cup behind.}
  \vspace{0.4\baselineskip}
  \label{cap_ex103} 
\end{subfigure}
\begin{subfigure}[b]{0.95\textwidth}
  \includegraphics[width=1\linewidth]{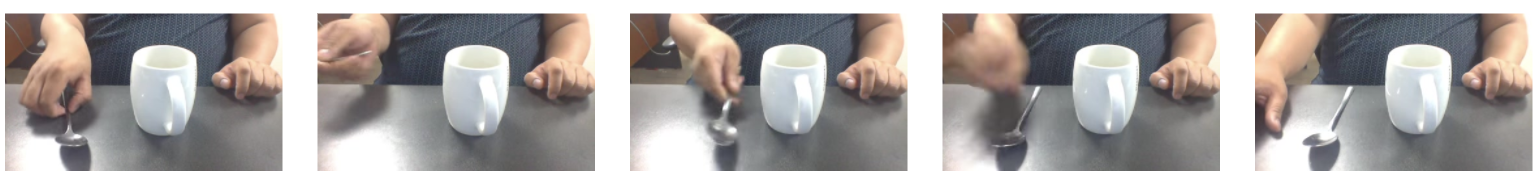}
  \vspace{-1.5\baselineskip}
  \caption{{\bf Ground Truth:} Hitting cup with spoon.}
  \vspace{-0.65\baselineskip}
  \caption{{\bf Model output:} Hitting mug with spoon.}
  \vspace{0.4\baselineskip}
  \label{cap_ex105} 
\end{subfigure}
\end{figure}

\begin{figure}[!h]
\begin{subfigure}[b]{0.95\textwidth}
  \includegraphics[width=1\linewidth]{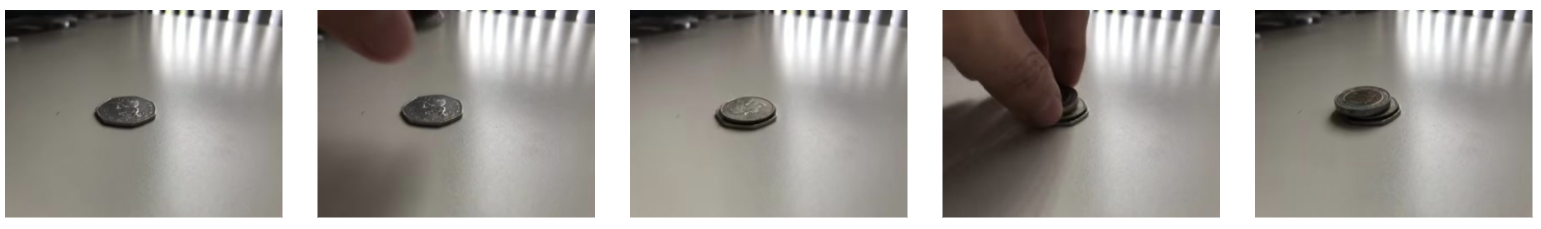}
  \vspace{-1.5\baselineskip}
  \caption{{\bf Ground Truth:} Stacking 4 coins. }
  \vspace{-0.65\baselineskip}
  \caption{{\bf Model output:} Piling coins up.}
  \vspace{0.4\baselineskip}
  \label{cap_ex106} 
\end{subfigure}
\begin{subfigure}[b]{0.95\textwidth}
  \includegraphics[width=1\linewidth]{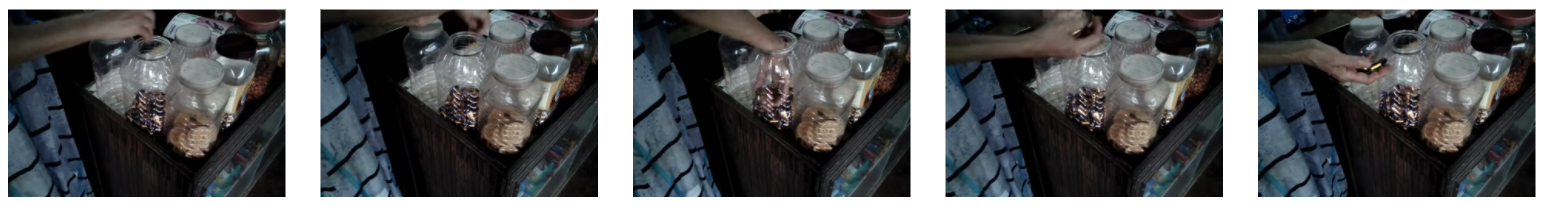}
  \vspace{-1.5\baselineskip}
  \caption{{\bf Ground Truth:} Taking toffee eclairs from jar.}
  \vspace{-0.65\baselineskip}
  \caption{{\bf Model output:} Taking battery out of container.}
  \vspace{0.4\baselineskip}
  \label{cap_ex107} 
\end{subfigure}
\begin{subfigure}[b]{0.95\textwidth}
  \includegraphics[width=1\linewidth]{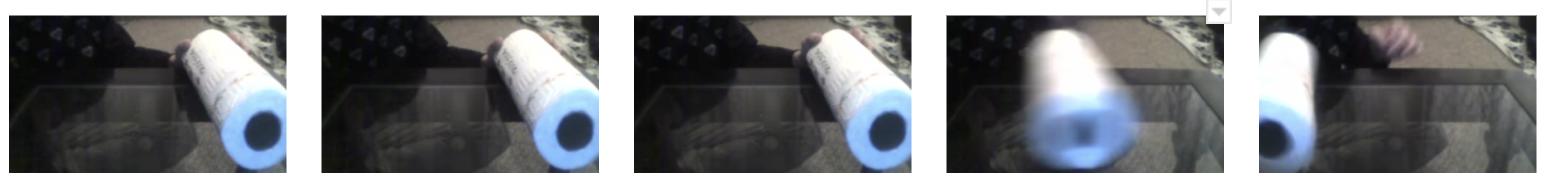}
  \vspace{-1.5\baselineskip}
  \caption{{\bf Ground Truth:} Rolling paper towels on a flat surface.}
  \vspace{-0.65\baselineskip}
  \caption{{\bf Model output:} Letting bottle roll along a flat surface.}
  \vspace{0.4\baselineskip}
  \label{cap_ex120} 
\end{subfigure}
\begin{subfigure}[b]{0.95\textwidth}
  \includegraphics[width=1\linewidth]{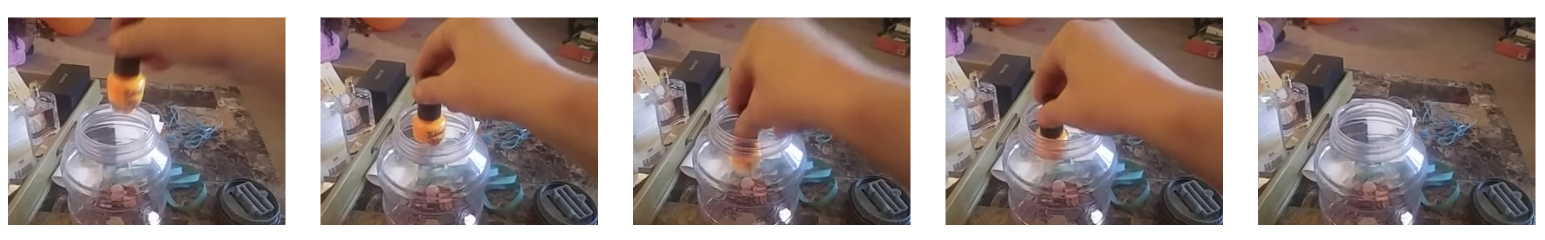}
  \vspace{-1.5\baselineskip}
  \caption{{\bf Ground Truth:} Pretending to put nail polish into jar.}
  \vspace{-0.65\baselineskip}
  \caption{{\bf Model output:} Pretending to put bottle into container.}
  \vspace{0.4\baselineskip}
  \label{cap_ex108} 
\end{subfigure}
\begin{subfigure}[b]{0.95\textwidth}
  \includegraphics[width=1\linewidth]{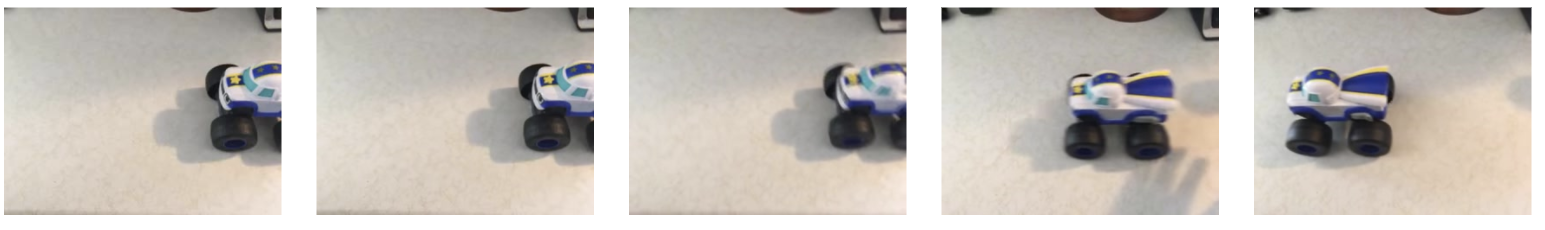}
  \vspace{-1.5\baselineskip}
  \caption{{\bf Ground Truth:} Letting toy truck roll along a flat surface.}
  \vspace{-0.65\baselineskip}
  \caption{{\bf Model output:} Pushing car from right to left.}
  \vspace{0.4\baselineskip}
  \label{cap_ex109} 
\end{subfigure}
\begin{subfigure}[b]{0.95\textwidth}
  \includegraphics[width=1\linewidth]{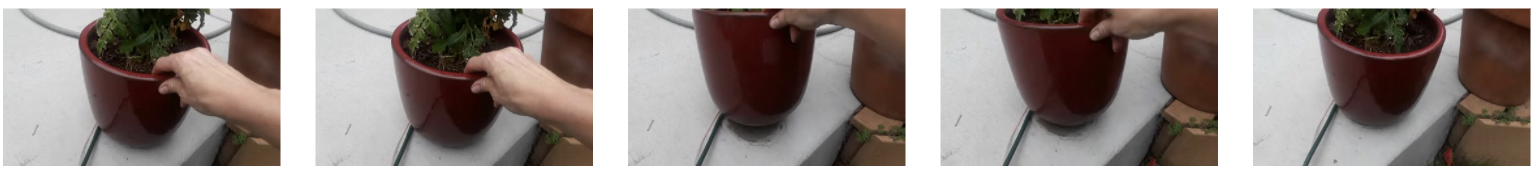}
  \vspace{-1.5\baselineskip}
  \caption{{\bf Ground Truth:} Lifting up one end of flower pot, then letting it drop down.}
  \vspace{-0.65\baselineskip}
  \caption{{\bf Model output:} Lifting up one end of bucket, then letting it drop down.}
  \vspace{0.4\baselineskip}
  \label{cap_ex110} 
\end{subfigure}
\begin{subfigure}[b]{0.95\textwidth}
  \includegraphics[width=1\linewidth]{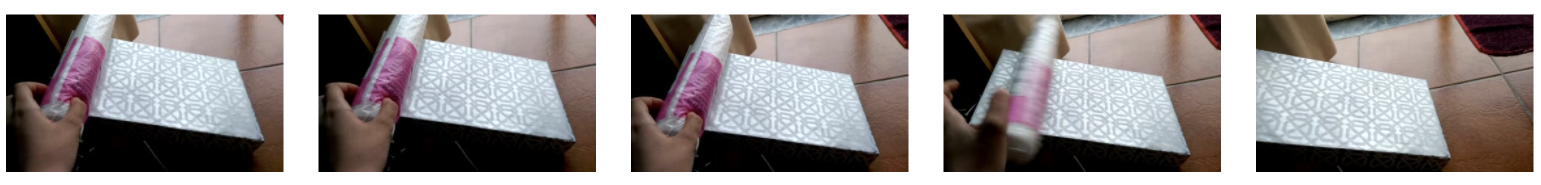}
  \vspace{-1.5\baselineskip}
  \caption{{\bf Ground Truth:} Letting roll roll down a slanted surface.}
  \vspace{-0.65\baselineskip}
  \caption{{\bf Model output:} Letting spray can roll down a slanted surface.}
  \vspace{0.4\baselineskip}
  \label{cap_ex112} 
\end{subfigure}
\begin{subfigure}[b]{0.95\textwidth}
  \includegraphics[width=1\linewidth]{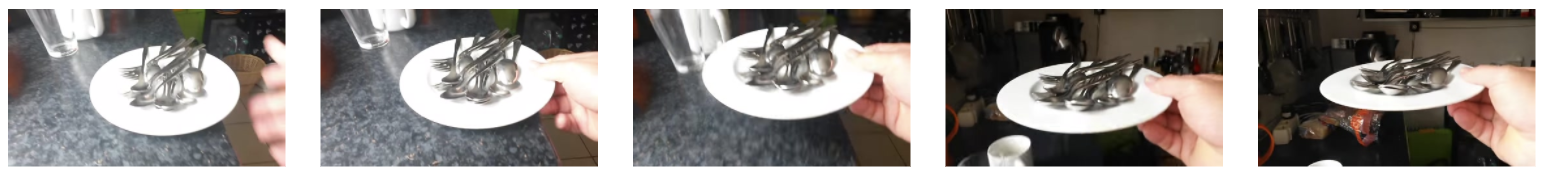}
  \vspace{-1.5\baselineskip}
  \caption{{\bf Ground Truth:} Lifting plate with cutlery on it.}
  \vspace{-0.65\baselineskip}
  \caption{{\bf Model output:} Lifting plate with spoon on it.}
  \vspace{0.4\baselineskip}
  \label{cap_ex114} 
\end{subfigure}
\end{figure}

\begin{figure}[!h]

\begin{subfigure}[b]{0.95\textwidth}
  \includegraphics[width=1\linewidth]{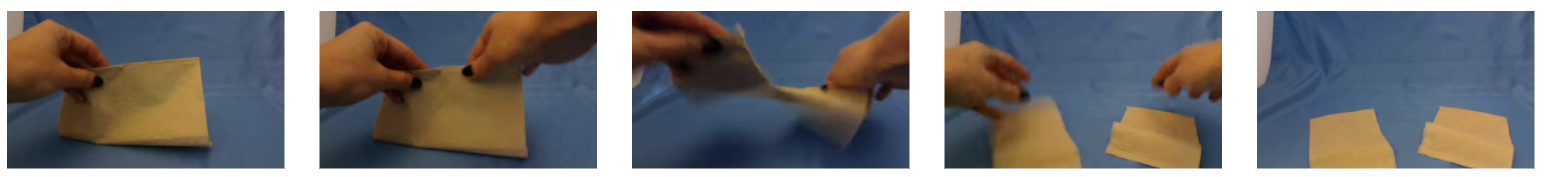}
  \vspace{-1.5\baselineskip}
  \caption{{\bf Ground Truth:} Tearing napkin into two pieces.}
  \vspace{-0.65\baselineskip}
  \caption{{\bf Model output:} Tearing paper into two pieces.}
  \vspace{0.4\baselineskip}
  \label{cap_ex115} 
\end{subfigure}
\begin{subfigure}[b]{0.95\textwidth}
  \includegraphics[width=1\linewidth]{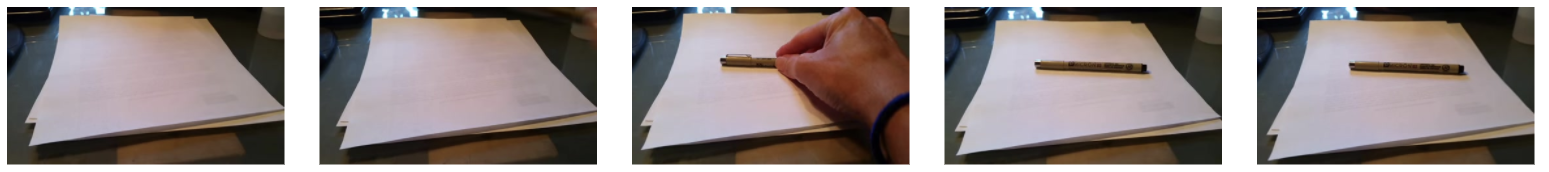}
  \vspace{-1.5\baselineskip}
  \caption{{\bf Ground Truth:} Putting pen on a surface.}
  \vspace{-0.65\baselineskip}
  \caption{{\bf Model output:} Putting pen that cant roll onto a slanted surface, so it stays.}
  \vspace{0.4\baselineskip}
  \label{cap_ex116} 
\end{subfigure}

\begin{subfigure}[b]{0.95\textwidth}
  \includegraphics[width=1\linewidth]{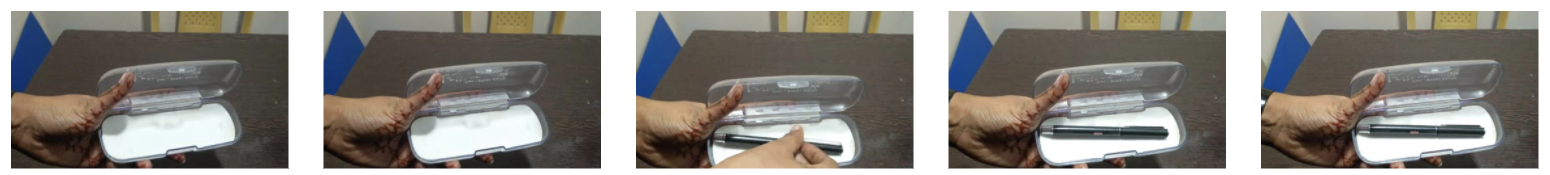}
  \vspace{-1.5\baselineskip}
  \caption{{\bf Ground Truth:} Putting pen into box.}
  \vspace{-0.65\baselineskip}
  \caption{{\bf Model output:} Showing that pen is inside the box.}
  \vspace{0.4\baselineskip}
  \label{cap_ex117} 
\end{subfigure}
\begin{subfigure}[b]{0.95\textwidth}
  \includegraphics[width=1\linewidth]{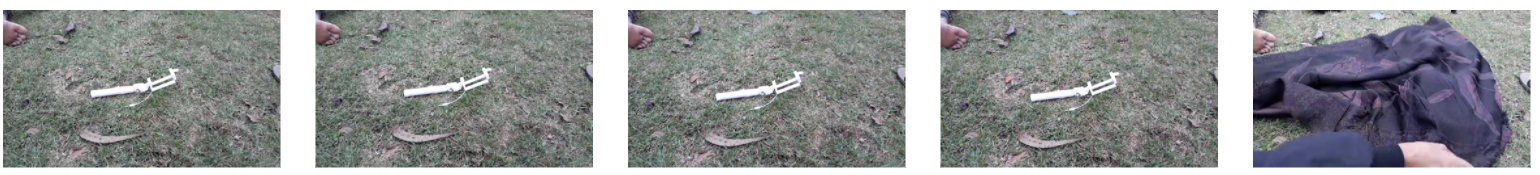}
  \vspace{-1.5\baselineskip}
  \caption{{\bf Ground Truth:} Covering selfi stick with shawl.}
  \vspace{-0.65\baselineskip}
  \caption{{\bf Model output:} Covering scissors with a blanket.}
  \vspace{0.4\baselineskip}
  \label{cap_ex118} 
\end{subfigure}
\begin{subfigure}[b]{0.95\textwidth}
  \includegraphics[width=1\linewidth]{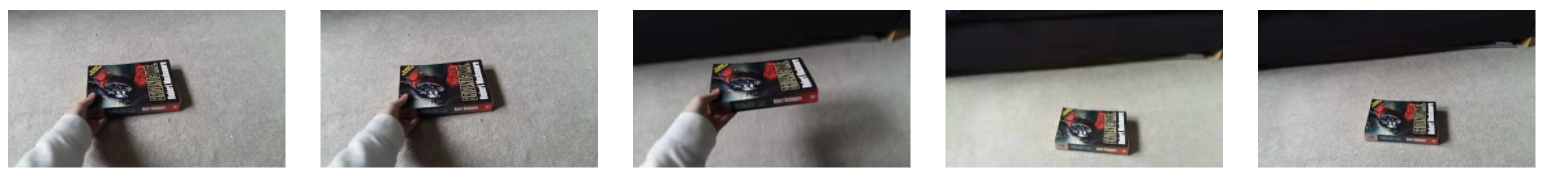}
  \vspace{-1.5\baselineskip}
  \caption{{\bf Ground Truth:} Lifting a book up completely, then letting it drop down.}
  \vspace{-0.65\baselineskip}
  \caption{{\bf Model output:} Lifting a book up completely, then letting it drop down.}
  \vspace{0.4\baselineskip}
  \label{cap_ex119} 
\end{subfigure}
\begin{subfigure}[b]{0.95\textwidth}
  \includegraphics[width=1\linewidth]{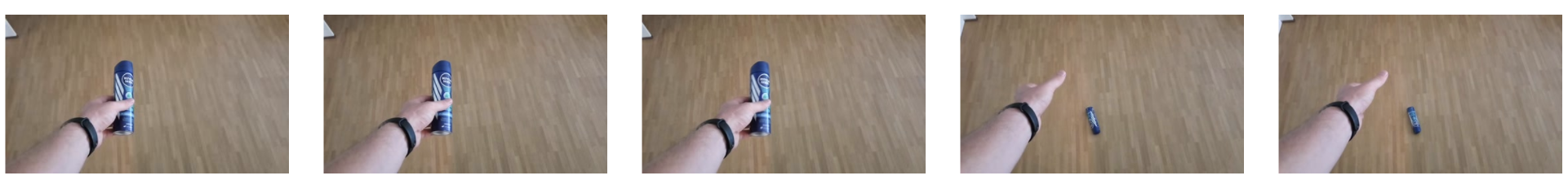}
  \vspace{-1.5\baselineskip}
  \caption{{\bf Ground Truth:} A deodorant falling like a rock.}
  \vspace{-0.65\baselineskip}
  \caption{{\bf Model output:} Bottle falling like a rock.}
  \vspace{0.4\baselineskip}
  \label{cap_ex3} 
\end{subfigure}
\begin{subfigure}[b]{0.95\textwidth}
  \includegraphics[width=1\linewidth]{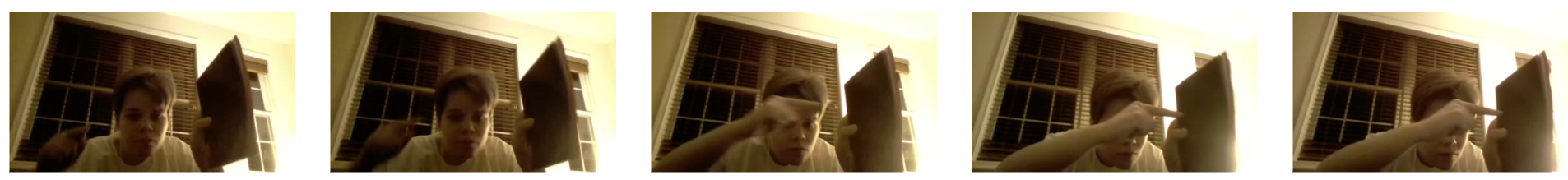}
  \vspace{-1.5\baselineskip}
  \caption{{\bf Ground Truth:}  Pretending to poke a book.}
  \vspace{-0.65\baselineskip}
  \caption{{\bf Model output:}  Pretending to poke a book. }
  \vspace{0.4\baselineskip}
  \label{captioning_examples}
\end{subfigure}
\caption{Ground truth captions and model outputs for video examples.}

\end{figure}

\clearpage

\subsection*{Visualization of classification model with Grad-CAM}

\begin{figure}[h!]
    \centering
    \includegraphics[width=\textwidth]{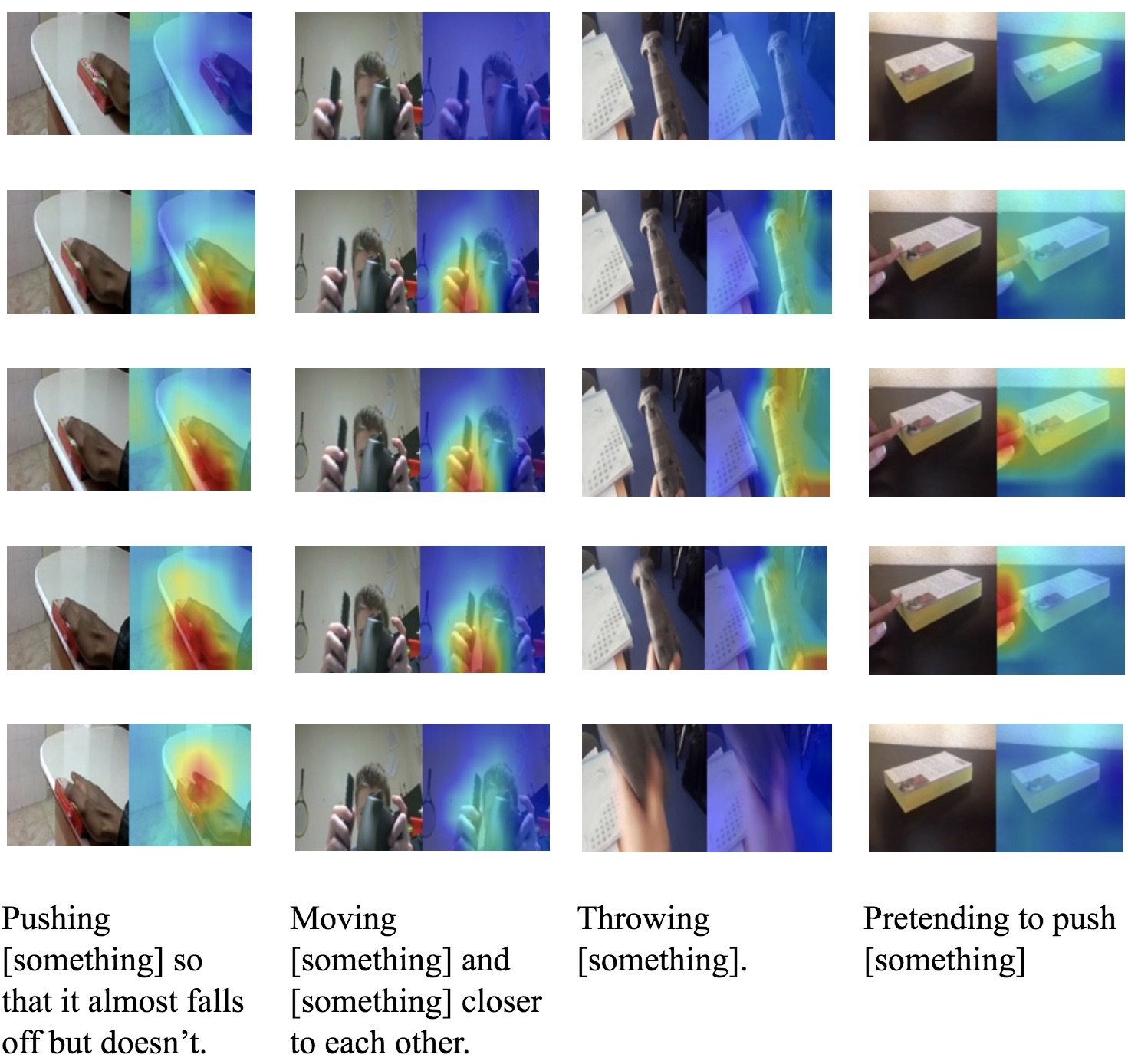}
    \caption{Grad-CAM for M($256$-$0$) on video examples predicted correctly during fine-grained action classification. We can see that the model focuses on different parts of different frames in the video in order to make a prediction.}
    \label{grad_cam_fig}
\end{figure}

To visualize regularities learned from data, we extracted temporally-sensitive saliency maps using Grad-CAM \cite{grad_cam_cite}, for both classification and captioning task. 
To this end we extended the Grad-CAM implementation for video processing. 
Figure \ref{grad_cam_fig} shows saliency maps of examples from Something-Something obtained with model M($256$-$0$) trained on fine-grained action categories, with $\lambda=1$ (i.e., the pure classification task).

\subsection*{Visualization of captioning model using Grad-CAM}

\begin{figure}[h!]
    \centering
    \includegraphics[width=\textwidth]{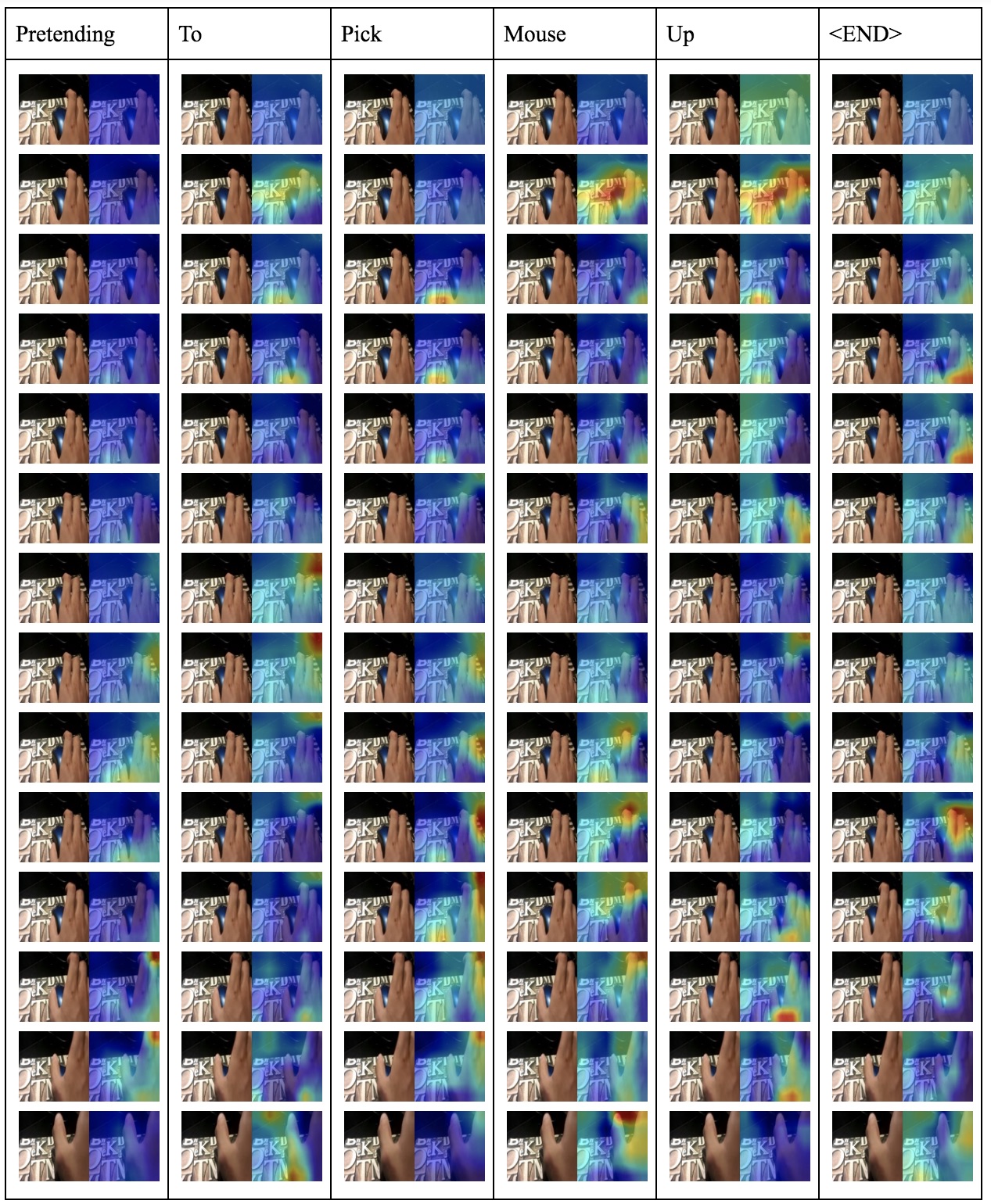}
    \caption{Grad-CAM on video example with ground truth caption {\bf Pretending to pick mouse up.} The model focuses on hand motion in the beginning and end of the video for the token ``Up''.}
    \label{grad_cam_cap}
\end{figure}

\begin{figure}[h!]
    \centering
    \includegraphics[width=\textwidth]{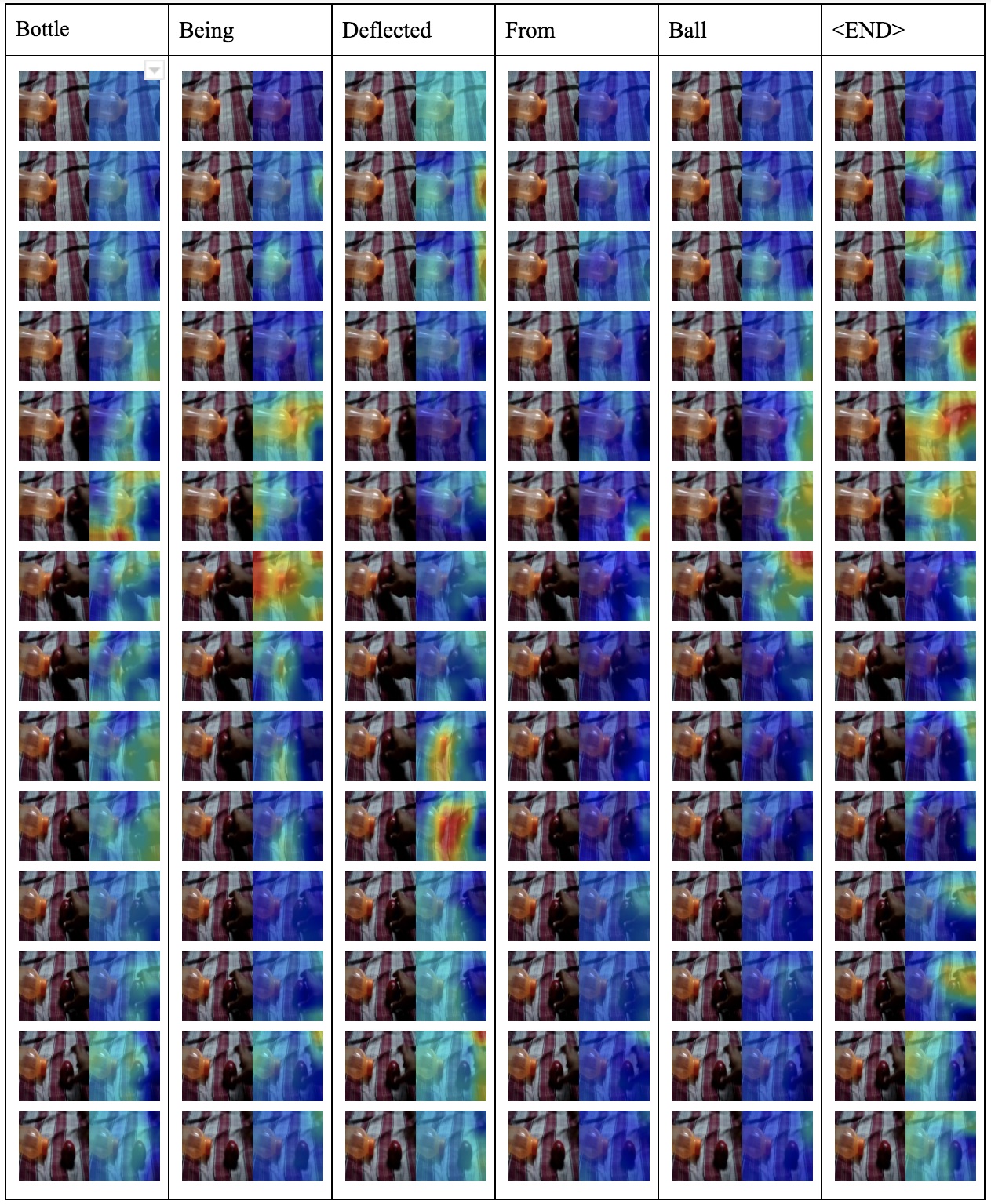}
    \caption{Grad-CAM on video example with ground truth caption {\bf Bottle being deflected from ball} during captioning process. The model focuses on the collision between bottle and ball, when using token ``Deflected''.}
    \label{grad_cam_cap2}
\end{figure}

To get saliency maps during the captioning process, we calculate the Grad-CAM once for each token, for which different regions of the video are highlighted.
Figures \ref{grad_cam_cap},-\ref{grad_cam_cap2} shows saliency maps for the captioning model, jointly trained with $\lambda=0.1$. 
Notice how the attentional focus of the model changes qualitatively as 
we perform Grad-CAM for different tokens in the target caption.
 
\vspace*{-0.7cm}

\begin{figure}[h!]
    \centering
    \includegraphics[width=\textwidth]{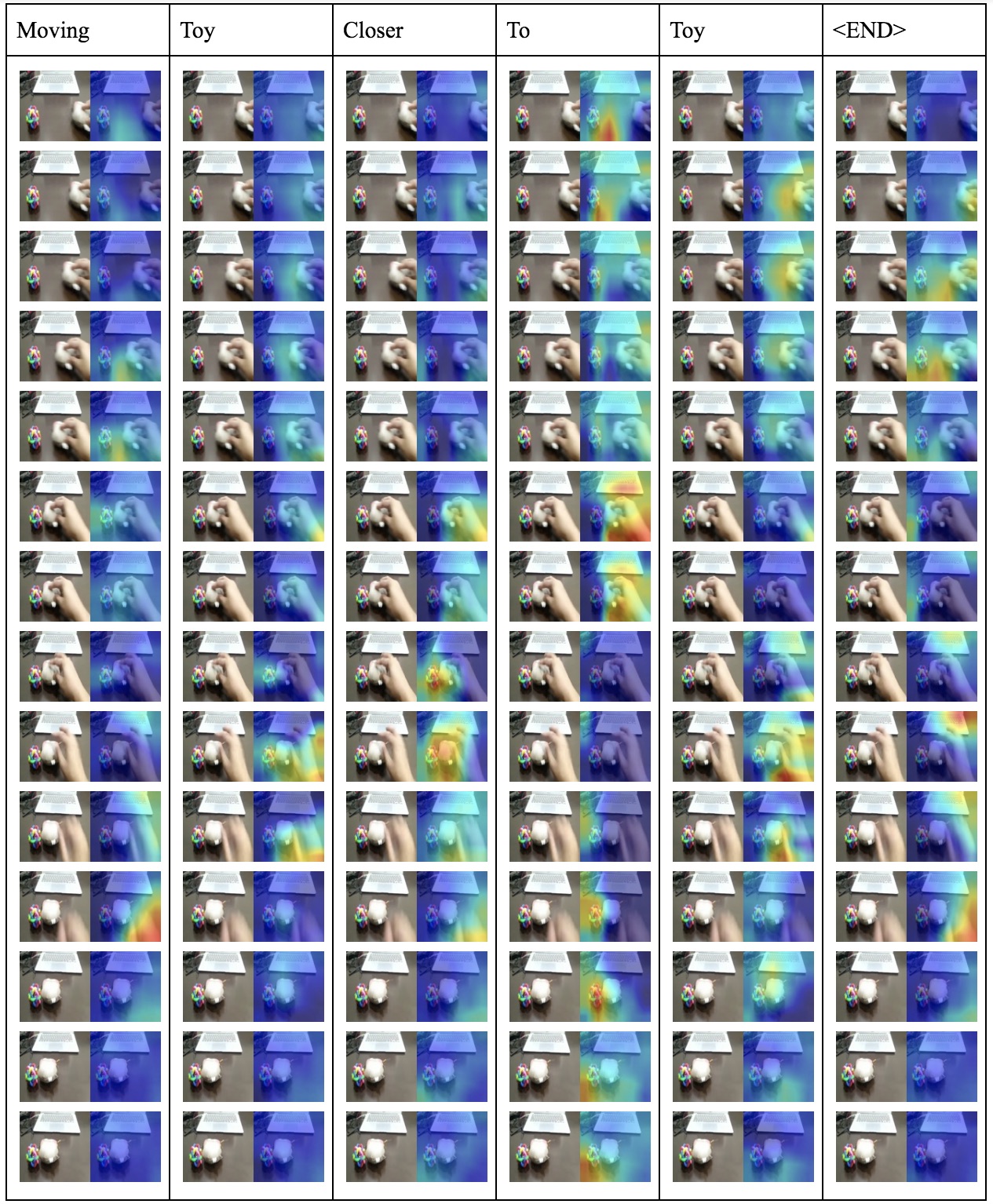}
    \caption{Grad-CAM on video example with ground truth caption {\bf Moving toy closer to toy.} We can see that the model focuses on the gap between toys when using ``Moving'' token. It also looks at both toy objects when using the token ``Closer''.}
    \label{grad_cam_cap1}
\end{figure}

\clearpage

\subsection*{20bn-kitchenware action categories}
Table \ref{kitchenware-labels} lists the full list of 20bn-kitchenware 
action categories. 

\begin{table}[h!]{\footnotesize}
\centering
\begin{tabular}{|l|}
\hline
\textbf{Action categories}                             \\ \hline
Using a fork to pick something up                                                              \\ \hline
Pretending to use a fork to pick something up                                                  \\ \hline
Trying but failing to pick something up with a fork                                            \\ \hline
Using a spoon to pick something up                                                             \\ \hline
Pretending to use a spoon to pick something up                                                 \\ \hline
Trying but failing to pick something up with a spoon                                           \\ \hline
Using a knife to cut something                                                                 \\ \hline
Pretending to use a knife to cut something                                                     \\ \hline
Trying but failing to cut something with a knife                                               \\ \hline
Using tongs to pick something up                                                               \\ \hline
Pretending to use tongs to pick something up                                                   \\ \hline
Trying but failing to pick something up with tongs                                             \\ \hline
Doing other things                                                                             \\ \hline
\end{tabular}

\caption{The 13 action categories represented in 20bn-kitchenware.}
\label{kitchenware-labels}
\end{table}

\end{document}